\documentclass[10pt,twocolumn,letterpaper]{article}

\usepackage[spanish,english]{babel}
\usepackage[utf8x]{inputenc}
\usepackage[T1]{fontenc}

\usepackage[a4paper,top=3cm,bottom=2cm,left=3cm,right=3cm,marginparwidth=1.75cm]{geometry}

\usepackage{amsmath}
\usepackage{graphicx}
\usepackage[colorinlistoftodos]{todonotes}
\usepackage[colorlinks=true, allcolors=blue]{hyperref}

\title{\textbf{Deep Generative Models for Drug Design and Response}}

\usepackage{authblk}
\author[]{Karina Zadorozhny}
\author[]{Lada Nuzhna}

\affil[]{\textit{Northwestern University}}
\date{}

\begin{document}
\maketitle
\selectlanguage{english}

\begin{abstract}
Designing new chemical compounds with desired pharmaceutical properties is a challenging task and takes years of development and testing. Still, a majority of new drugs fail to prove efficient \textit{in vivo}. Recent success of deep generative modeling holds promises of generation and optimization of new molecules. In this review paper, we provide an overview of the current generative models, and describe necessary biological and chemical terminology, including molecular representations needed to understand the field of drug design and drug response. We present commonly used chemical and biological databases, and tools for generative modeling. Finally, we summarize the current state of generative modeling for drug design and drug response prediction, highlighting the state-of-art approaches and limitations the field is currently facing.
\end{abstract}

\section{Introduction}
The path from initial screening to marketable drugs takes around 20 years \cite{paul_how_2010}, with cost being anywhere from \$0.5 to \$2.6 billion dollars \cite{avorn_26_2015}. Given the remarkable impediments associated with drug design, it is necessary to develop more precise strategies to propose drug candidates. Computational chemistry approaches to drug design, often called computer-aided drug design, received a lot of attention in the past 30 years. The growing availability of high-throughput screening (HTS) data holds a great potential to accelerate the development of efficient new compounds.

Pharmacologically active drug compounds act by binding to their target (which is most commonly a protein) and altering their function by either stimulating or inhibiting it. Broadly speaking, the current drug design approaches can be divided into two categories. The first one is ligand-based (or indirect) approaches that rely on existing molecules known to be a binding partner of some target and try to mimic their properties. These approaches include molecular fingerprint similarity search, pharmacophore modeling, or quantitative structure-activity relationship (QSAR). The second approach is structure-based (or direct) where a 3D structure of a desired target (or a target that is highly similar to the desired one) is known. New molecules are then designed to bind to the target based on its structural properties. The methods that fall in this category include molecular docking or molecular dynamic simulations.

Deep generative models (DGM) have attracted a lot of attention in recent years. Given their success in image and text generation \cite{guzel_turhan_recent_2018, aggarwal_generative_2021-1}, there is an ongoing effort to apply these approaches to the medical field as well. Current applications include medical image analysis, data augmentation, generation of new DNA, RNA, or protein sequences with specific properties and missing regions in known protein structures \cite{lan_generative_2020-1}.

Despite the optimism, designing novel drugs is particularly difficult due to the graphical structure of molecules, chemical validity, and feasibility of synthesis of generated molecules. Several major simplifications have been used so far to reduce the complexity of the task. For example, instead of working with and generating full molecular structures, molecules can be described by their properties in binary strings. However, to fully automate the drug design process, it is necessary to incorporate complex information, such as chirality and structural conformations, into the molecular representations.

Strikingly, 75-97\% of novel drugs fail human clinical trials. The primary reason for failing a trial is a lack of demonstrated efficiency and the second most common reason is safety concerns \cite{wong_estimation_2019}.  Therefore, optimizing for efficient physiological response and low toxicity in addition to favorable chemical properties is at the core of the approaches that could lead to successful drug design. This is encompassed in the field of drug response prediction.  

In this review article, we describe the current DGM approaches for drug design with the aim of bridging the fields of drug design and drug response prediction to allow for designing more efficient and more powerful pipelines. We first briefly introduce the most common neural architectures that are used for drug design and drug response prediction. We also list the most common databases and tools, and compare current molecular data representations. We describe state-of-the-art techniques for small molecule design, gene therapy, and protein sequence generation. Finally, we conclude our review with remarks on the most challenging aspects of drug design and considerations about the direction of future research. 

\begin{figure*}[h!]
    \centering
    \includegraphics[width=2\columnwidth]{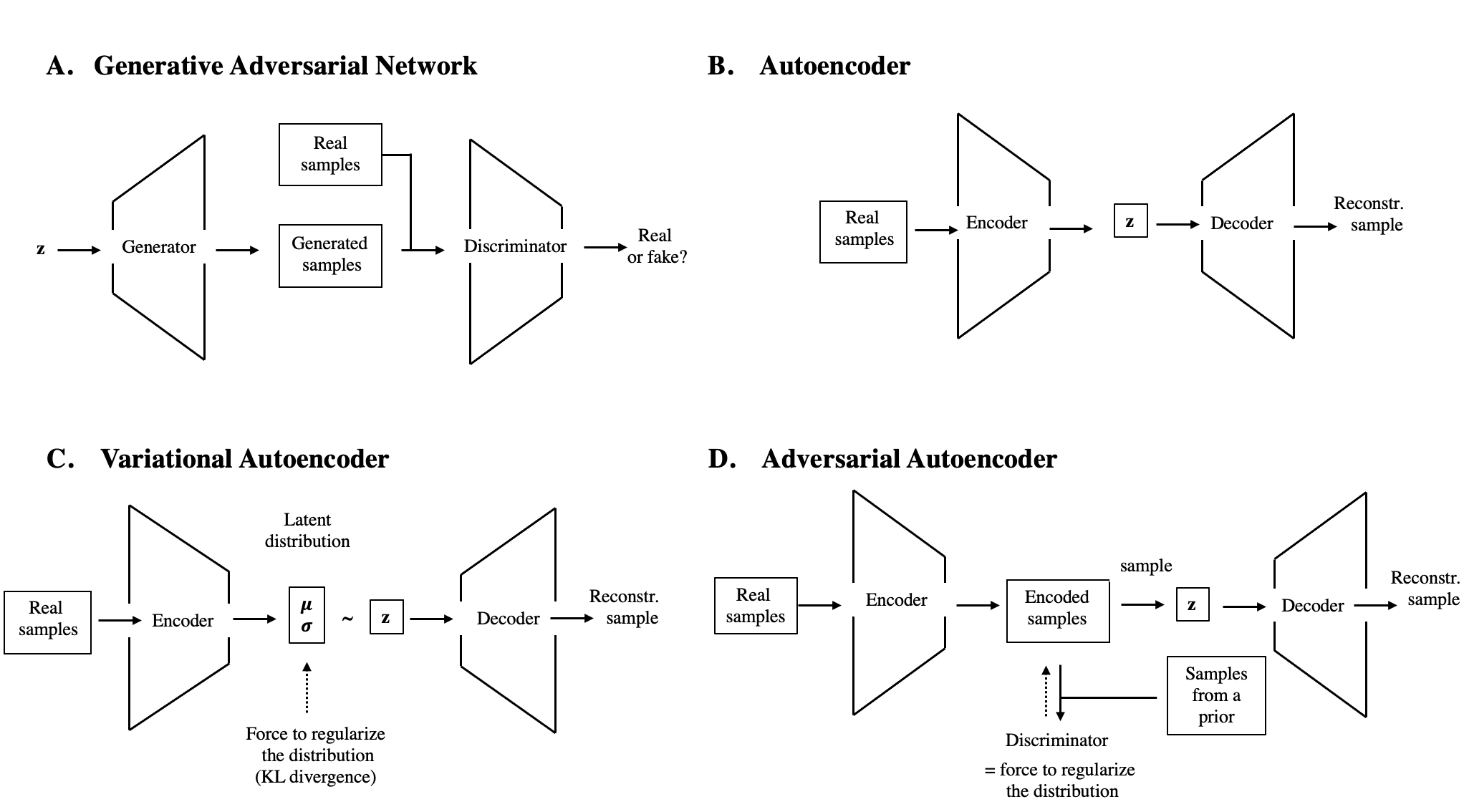}
    \caption{Generative Models. A. Generative Adversarial Network (GAN). B. Autoencoder. C. Variational Autoencoder (VAE). D. Adversarial Autoencoder (AAE).}
    \label{fig:models}
\end{figure*}

\section{Overview of generative models}
Deep generative models have achieved great success ranging from generation of high-quality images to large corpora of meaningful text.  Formally, deep generative models are neural networks trained to model a distribution of training samples and are able to sample from the learned distribution to generate new samples.  Explicit generative models specify the full data distribution and allow to express likelihoods that a given sample comes from the distribution. On the other hand, implicit generative models only specify a procedure of drawing samples to generate new data but do not provide a way to express a full distribution.

\subsection{GANs}
Generative Adversarial Networks (GANs) are implicit density models introduced in 2014 \cite{goodfellow_generative_2014}. The idea behind generating samples with GANs is to have two networks - one is a generator model that creates new samples from a random vector and the other is the discriminator (or the critic) that aims to distinguish newly generated samples from the real ones. These two parts are competing with  each other. The generator gets better at generating realistically looking samples while the discriminator gets better at classifying those that are real and those that are fake. After both networks have been trained, only the generator is used to create realistic inputs from a random starter vector.   

\subsection{Autoencoders}
A standard autoencoder (AE) has two parts - an encoder and a decoder. The inputs are passed through an encoder that outputs some representation of these inputs to the latent space. From there, the decoder can reconstruct the representation back such that it is as close as possible to the original input. Most commonly, the latent space is chosen such that it has fewer dimensions than the inputs. This leads to compression of the samples and extraction of the most useful information. The encoded latent space, which contains all the information that is necessary to retrieve the original input, can then be used for downstream tasks. For example, a classifier can be fitted on top of a learned latent space. Compared to Principal Component Analysis (PCA), autoencoders can perform non-linear dimensionality reduction. Autoencoders can also be used for denoising where a noisy input is reconstructed without the unwanted noise.
 
\subsection{VAEs}
Variational Autoencoders (VAEs) are regularized probabilistic autoencoders \cite{kingma_auto-encoding_2014}. Similar to a standard autoencoder, they contain an encoder and a decoder. However, instead of encoding each sample to some point a latent space, the encoder defines a whole distribution. From this distribution, vectors are being sampled and fed into a decoder that reconstructs them back to be as close as possible to the original input. To give the latent space distribution some specific properties, it is regularized during training. Most commonly, we force the latent space to follow a normal distribution since that provides a closed form for many equations and is simple to work with. The loss function for VAEs thus contains two terms – a reconstruction error and a regularization term which is a Kullback-Leibler divergence between the learned latent distribution and a specified prior (normal distribution).
 
\subsection{AAEs} 
Adversarial Autoencoders (AAEs) are regularized autoencoders that adapt GANs training technique \cite{makhzani_adversarial_2016}. Similar to VAEs, they impose certain properties on the learned latent space such as being normally distributed but they use a different regularization technique. Apart from an encoder and a decoder, there is an auxiliary adversary network which is trained to distinguish samples from the latent space and samples drawn from a chosen distribution. The encoder tries to fool the adversary by producing samples that are indistinguishable from the chosen distribution and this serves as regularization of the latent space. One of the benefits of the AAEs compared to VAEs is that the chosen prior distribution that the encoder tries to mimic can be any distribution, even a non-parametric one. Depending on the choice of the encoder, the AAE can be either a deterministic model that maps all the inputs to the same space in the latent space or have a probabilistic encoder that outputs parameters for  a full latent posterior distribution.

\subsection{Transformers} 
Transformer is an attention-based architecture that gained popularity in recent years due to its successful use in language models \cite{devlin_bert_2019, brown_language_2020}. Transformer has two functional parts: a block of stacked encoders and a block of decoders (Figure 2). Encoding part produces a representation of the input that tells us how different parts of the input sequence are related to each other. “Relevancy” is determined by the attention formula.  This approach allows us to establish connections between parts of the sequence that can be far from each other. This makes it attractive for use in protein structure prediction, where functional groups at different sides of the sequence can interact with each other, dictating the 3D conformation of the sequence. 

\begin{figure}
    \centering
    \includegraphics[width=\columnwidth]{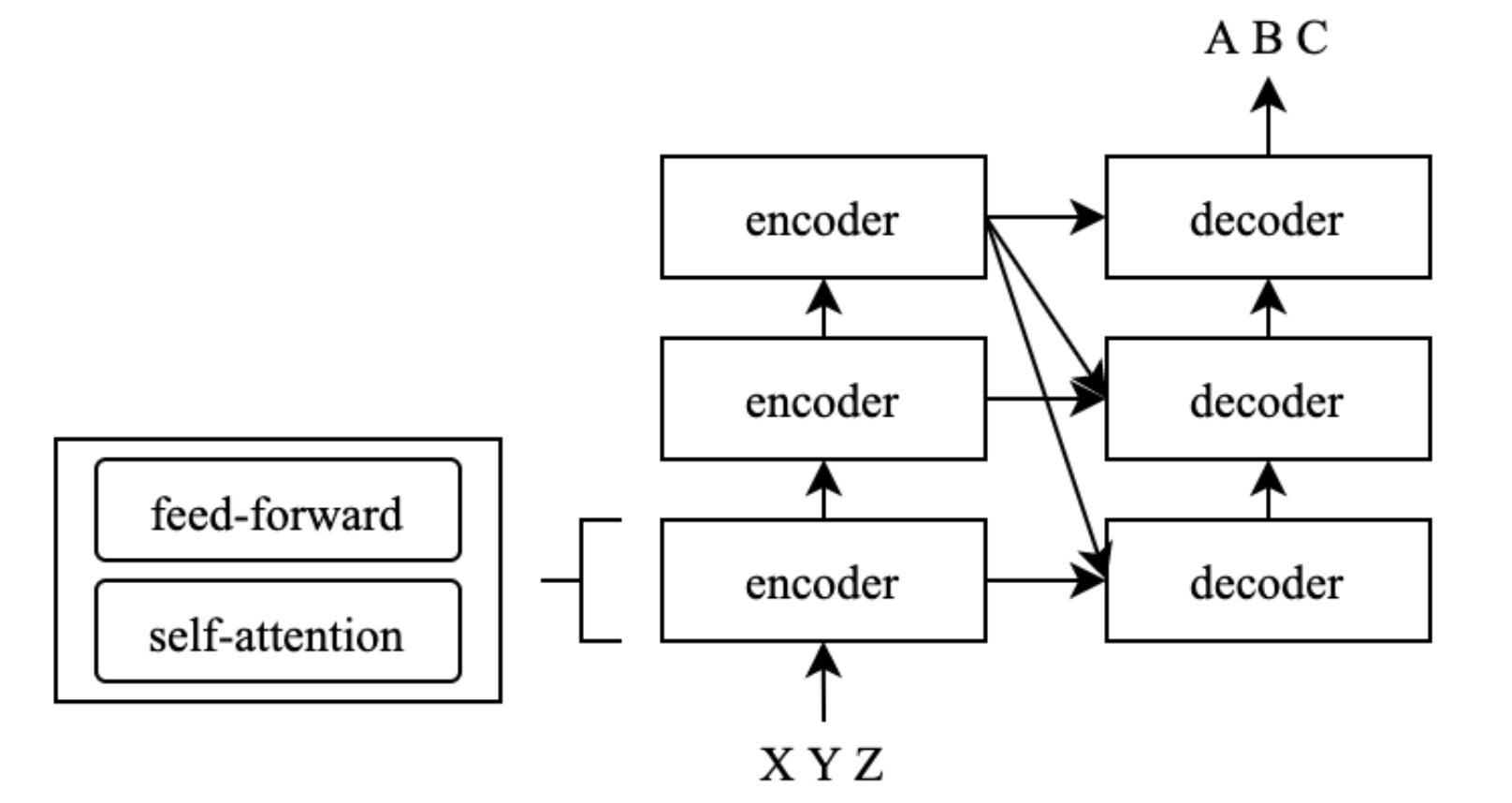}
    \caption{Transformer architecture.}
    \label{fig:transf}
\end{figure}

\section{Databases and tools}
In this section, we highlight some of the most useful databases and tools for drug design and drug response prediction.

\subsection{Databases of Chemical Compounds}
Table \ref{tab:chem} shows the most commonly used databases of binding affinities, bioactivity, and 3D structures of chemical compounds.

\begin{table*}
\centering
\scalebox{0.75}{
    
    \begin{tabular}{|l|l|l|l|l|}
    \hline
    \textbf{Database} & \textbf{Type} & \textbf{Note}                           & \textbf{Entries} & \textbf{Link}                \\ \hline
    BindingDB &
      Binding affinity &
      \begin{tabular}[c]{@{}l@{}}Database containing binding affinities \\ between proteins and ligands with \\ drug-like properties.\end{tabular} &
      1.5 M &
      \begin{tabular}[c]{@{}l@{}}\href{BindingDB.org}{BindingDB.org}\end{tabular} \\ \hline
    ZINC &
      Bioactivity, 3D structure &
      Commercially available molecules &
      21 M &
      \begin{tabular}[c]{@{}l@{}}\href{https://zinc.docking.org/}{zinc.docking.org}\end{tabular} \\ \hline
    PubChem &
      Bioactivity, 3D structure &
      \begin{tabular}[c]{@{}l@{}}Contains compounds,   \\ substances (mixtures,  extracts), \\ and high-throughput screenings (BioAssay)\end{tabular} &
      40 M &
       \begin{tabular}[c]{@{}l@{}}\href{https://pubchem.ncbi.nlm.nih.gov/}{pubchem.ncbi.nlm.nih.gov}\end{tabular} \\ \hline
    PDB &
      3D structure &
      \begin{tabular}[c]{@{}l@{}}Contains 3D structures of proteins \\ and nucleic acids.\end{tabular} &
      170 000 &
      \begin{tabular}[c]{@{}l@{}}\href{https://www.rcsb.org/}{rcsb.org}\end{tabular} \\ \hline
    PDBeChem          & 3D structure  & Ligands referenced in the PDB database. & 14 500           &  \href{https://www.ebi.ac.uk/pdbe-srv/pdbechem/}{ebi.ac.uk}   \\ \hline
    ChemDB            & 3D structure  & Commercially available molecules.       & 5 M              &  \href{http://chemdb.ics.uci.edu/}{chemdb.ics.uci.edu}     \\ \hline
    ChemSpider        & 3D structure  & Crowdsourced chemical database.         & 100 M            &  \href{http://www.chemspider.com/}{chemspider.com} \\ \hline
    \end{tabular}
    }
    \caption{Databases of chemical compounds.}
    \label{tab:chem}
\end{table*}

\begin{table*}
\centering
\scalebox{0.75}{
    \begin{tabular}{|l|l|l|l|l|}
    \hline
    \textbf{Database} &
      \textbf{Type} &
      \textbf{Note} &
      \textbf{Size} &
      \textbf{Link} \\ \hline
    NCI-60 &
      Drug response &
      \begin{tabular}[c]{@{}l@{}}Cytotoxic and cytostatic effects of different \\ drugs on human tumor   cell lines.\end{tabular} &
      \begin{tabular}[c]{@{}l@{}}60 cell lines\\  with 100 000 drugs\end{tabular} &
       \href{https://dtp.cancer.gov/discovery_development/nci-60/}{dtp.cancer.gov} \\ \hline
    CTRPv2 &
      Drug response &
      \begin{tabular}[c]{@{}l@{}}Cytotoxic and cytostatic effects of different\\  drugs on human tumor cell lines.\\  Provides different concentrations of drugs.\end{tabular} &
      \begin{tabular}[c]{@{}l@{}}850 cell lines\\  with 481 drugs\end{tabular} &
       \href{https://portals.broadinstitute.org/ctrp.v2.1/}{broadinstitute.org} \\ \hline
    GDSC &
      Drug response &
      \begin{tabular}[c]{@{}l@{}}Cytotoxic and cytostatic effects of different \\ drugs on human tumor cell lines.\end{tabular} &
      \begin{tabular}[c]{@{}l@{}}700 cell lines \\ with 138 drugs\end{tabular} &
      \href{https://www.cancerrxgene.org/}{cancerrxgene.org}\\ \hline
    CCLE &
      Drug response &
      \begin{tabular}[c]{@{}l@{}}Cytotoxic and cytostatic effects of different\\  drugs on human tumor cell lines.\end{tabular} &
      \begin{tabular}[c]{@{}l@{}}1000 cell lines \\ with 24 drugs\end{tabular} &
      \href{https://ccle.ucla.edu/}{ccle.ucla.edu} \\ \hline
    NCBI GEO &
      \begin{tabular}[c]{@{}l@{}}Gene expression\\ profiles\end{tabular} &
      \begin{tabular}[c]{@{}l@{}}Gene expression data from human and \\ animal tissues.\end{tabular} &
      – &
     \href{https://www.ncbi.nlm.nih.gov/geo/}{ncbi.nlm.nih.gov} \\ \hline
    CMap-L1000v1 &
      \begin{tabular}[c]{@{}l@{}}Gene expression\\ changes\end{tabular} &
      \begin{tabular}[c]{@{}l@{}}Expression changes (perturbations)\\  of 1000 landmark genes for different\\  concentrations and durations of drug treatments.\end{tabular} &
      \begin{tabular}[c]{@{}l@{}}1000 genes \\ with 2700 drugs\end{tabular} &
       \href{https://cmap.ihmc.us/}{cmap.ihmc.us} \\ \hline
    \end{tabular}
    }

\caption{Databases of cell line and tissue screenings.}
\label{tab:cell_lines}
\end{table*}

\subsection{Databases of Cell Line Screens}
In Table \ref{tab:cell_lines}, we present open source databases of drug response (sensitivity) data on cancer cell lines as well as gene expression data collected using high-throughput screenings for cells and tissues. 

\subsection{Tools}
QSAR is a computational method for modeling relationships between structural properties of chemical compounds and their biological properties (such as enzyme activity, minimum effective dose, toxicity). This can often be used to validate the results achieved by deep nets and for benchmarking. However, QSAR is sometimes at risk of giving inaccurate predictions.

RDKit open-source chemoinformatics library (available from http://www.rdkit.org) can be used for a number of 2D and 3D molecular operations, from stereochemistry identification to detecting chemistry problems. 

\section{Data Representations}
Given the diversity of problems and screenings that drug discovery can cover, data representations also differ significantly. We present some of the most common ones. For a detailed review of the molecular representations, see David et al. \cite{david_molecular_2020}. We stress that there exist  molecules that cannot be described using a standard notation.

\subsection{SMILES and Its Variants}

Simplified Molecular Input Line Entry System (SMILES) represents molecules as strings and, to date, is one of the most commonly used molecular representations. 
The strings are created by traversing a molecular graph in a specific order (for example, RDKit uses a depth-first search algorithm). SMILES strings are not unique and one molecule can have several representations. Therefore, a canonical string is often chosen for a given molecule. 
 
SMILES Arbitrary Target Specification (SMARTS) is an extension of SMILES that allows to encode more information about bond types, connectivity, and atomic properties in a string. 

Another string representation, called SMIRKS allows for expressing reactants and products of a generic chemical reaction.
 
\subsection{Molecular Fingerprints}
Molecular fingerprints are another popular way of encoding molecular structures. In its simplest form, binary encoding indicates presence of absence of individual substructures within a molecule. However, any information about bonds and other properties is necessarily lost with this representation and it is not a one-to-one mapping.

Circular fingerprint considers connectivity of atoms up to a certain distance to solve the problem of non-uniqueness of many representations. Every atom receives a descriptor that indicates its closest neighbors. Examples of circular fingerprints include extended-connectivity fingerprints (ECFPs) such as ECFP4 or ECFP6 where the number indicates the radius around an atom to be considered. Other examples are Multilevel Neighborhoods of Atoms (MNA) or MolPrint2D Format (MPD).
 
A fingerprint that describes molecules' surface properties are called spectophores. Molecules that have similar shapes and 3D properties will retains a similar representations.

\subsection{Peptide and Protein Representations}
The simplest way of representing peptides or proteins is using amino acid sequences. Each individual amino acid, a building blocks of peptides, is assigned a letter and the final peptide is represented by a string of these letters.

There are other representations of proteins such as distance and contact maps that preserve information about protein structures. Protein contact maps represent distances between all pairs of amino acids in a 3D protein structure in a matrix.

\subsection{Graph Representations}
Molecular graphs can be also described directly using graphical representations.

Standard adjacency matrices can be used to represent a presence or absence of bonds between atoms in a graph.

A node feature matrix is a larger matrix can be used to indicate atom types, formal charges, or other properties of each atom.

Similarly, in an edge feature matrix, each row represents an edge in a graph and is a binary vector that indicates properties of that such (such as the type of a bond).

\begin{figure}
    \centering
    \includegraphics[width=\columnwidth]{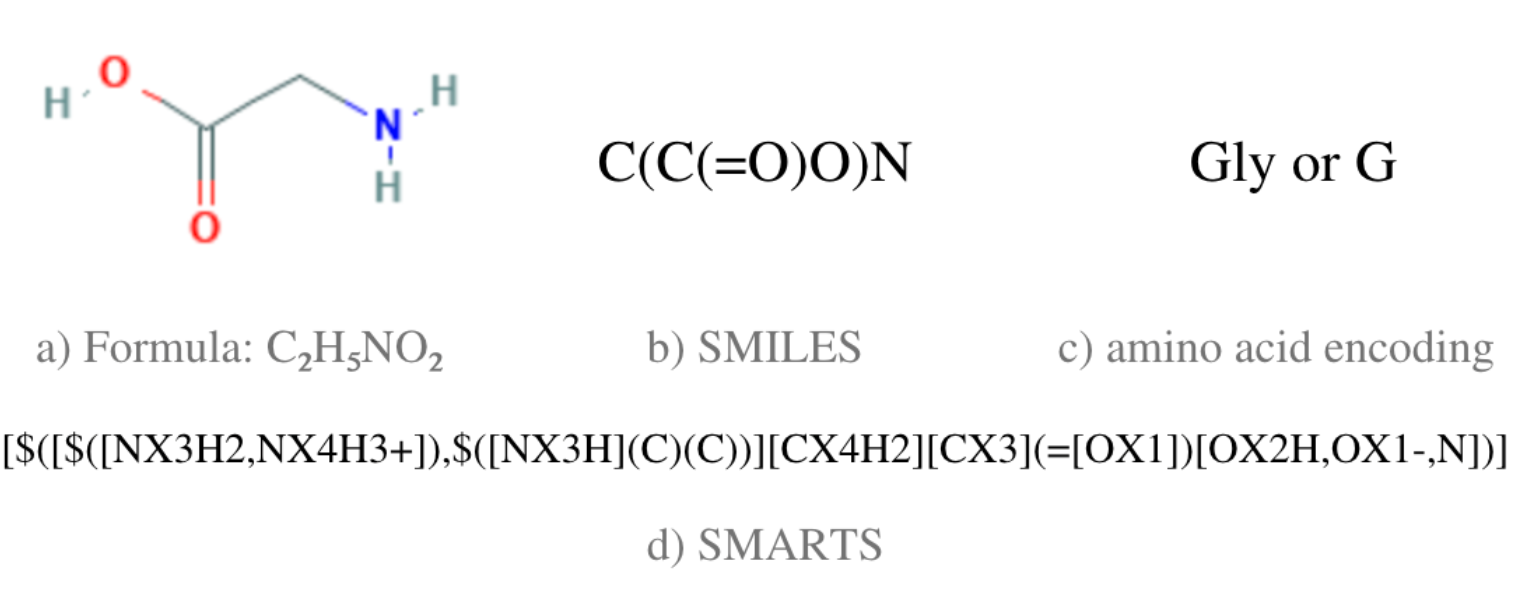}
    \caption{Example of molecular representations. Amino acid glycine represented in different formats.}
    \label{fig:repres}
\end{figure}

\section{Drug Design}
 \subsection{Small Molecule Design}
With many variations of generative architectures published in recent years, there are a number of ways to enforce properties onto generated structures. Desired drug features, such as growth inhibition of tumors (or GI50), can be used as input to encoder in Conditional Autoencoders \cite{joo_generative_2020}. Some work was also done to frame drug design as a ‘translation’ problem, applying transformer architecture for converting amino-acid sequence to SMILES representation of the molecule that can bind it \cite{grechishnikova_transformer_2020}. However, even with faster screening methods, docking validation still remains a bottleneck in this type of candidate generation. 
Moreover, when mapping discrete representations into continuous space, “grammar” of chemical  syntax is hard to preserve. This can lead to unrealistic aromatic systems and incorrect atomic valences. Because of this, SMILES-generated molecules often require additional “correcting” architectures. 

One possible approach for correction is sequence-to-sequence learning with attention, where the input is a wrong sequence and the target is the corrected one \cite{druchok_towards_2019}. Grammar VAE (GVAE) \cite{kusner_grammar_2017-1} was also used to generate SMILES that follow syntactic constraints given by a context-free grammar. Finally, syntax-directed VAE (SDVAE) \cite{dai_syntax-directed_2018}) can make use of attribute grammar to enforce syntactic and semantic constraints on generated SMILES.

\begin{figure}[h!]
    \centering
    \includegraphics[width=\columnwidth]{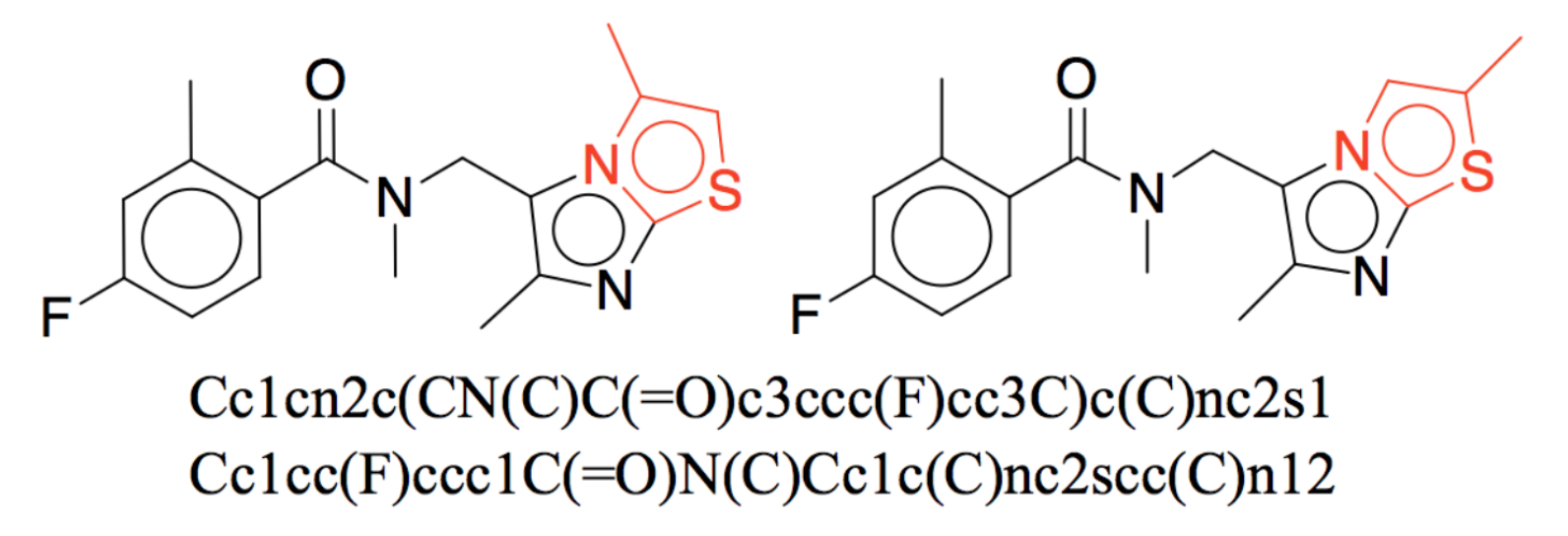}
    \caption{SMILES strings for these two very similar molecules show only 50.5\% similarity.  Adapted from \cite{jin_junction_2019} .}
    \label{fig:smiles_sim}
\end{figure}

While focusing on SMILES strings or molecular fingerprints simplifies generation of new molecules, a lot of information is lost using these approaches. For instance, chemically similar molecules can have very different SMILES strings (Figure \ref{fig:smiles_sim}. This representation prevents learning meaningful smooth embedding.

Graph neural networks that capture graph structures by message passing between the nodes are attractive methods for generation of full molecular structures \cite{jiang_could_2021}. Given that graphs are discrete structures with arbitrary connectivity, graph generation is a difficult domain and relatively young domain of generative modeling.

One approach is to generate graph nodes and edges one by one. This was done using an LSTM-based autoregressive model \cite{li_learning_2018}.  Simonovsky and Komodakis, \cite{simonovsky_graphvae_2018} used GraphVAE to directly generate nodes (atoms) and predict an adjacency matrix indicating edges (bonds) of a graph. Another approach, which enhances chemical validity of generated molecular graphs, is to use smaller subgraphs as building blocks of a final molecular graph. Jin et al., \cite{jin_junction_2019} used a model called Junction Tree VAE (JT-VAE), which first generated a junction tree that serves as a scaffold for the molecule and then specifies the subgraphs as nodes of the junction tree. Their method of coarse-to-fine generation of molecules significantly outperformed previous SMILES-based approaches such as GrammarVAE \cite{kusner_grammar_2017-1}, Character VAE \cite{gomez-bombarelli_automatic_2018-1}, or graph generating GraphVAE in validity of generated molecules. Another model , DeLinker, which is a gated graph neural network (GGNN) that links two molecular fragments together to form more complex compounds \cite{imrie_deep_2020} also incorporates 3D structural information to constrain the linking process. However, DeLinker’s performance was not compared to previous methods.

\subsection{Gene Therapy}
Generative models have a particular promise for the area of gene therapy, where adeno-associated virus (AAV) capsids serve as vectors for gene delivery. However, due to wide-spread immunity against common AAVs, capsids often need to be diversified such that antigens on their surface aren't recognized by the human immune system. 
Previous approaches, like directed evolution or DNA shuffling between serotypes, often produced unstable results (Bryant et al., 2021; Ogden et al., 2019). Earlier this year Dyno Therapeutics used variational autoencoders to create novel and stable adeno-associated viral capsids \cite{sinai_generative_2021}. This case also emphasizes that, compared to supervised methods, unsupervised generation can potentially take advantage of evolutionary information.

\subsection{Proteins and Peptides}
Peptides are short sequences of amino acids that can be used as therapeutics. There are over 60 peptides approved for clinical use by the FDA. While many naturally-occurring peptides are approved for medical use, they often end up having some weaknesses, such as low membrane permeability, chemical instability and short half-life. Synthetic peptides generated by computational methods can overcome these weaknesses \cite{fosgerau_peptide_2014}.

Das et al. \cite{das_accelerated_2021}, compared usage of Wasserstein Autoencoder (use Mean Maximum Discrepancy as a regularization in their loss) and Beta Variational Autoencoders (Kullback-Leibler divergence as regularization) for the screening of peptides with antimicrobial properties. WAE managed to learn meaningful latent representation that captured correlation between evolutionary similarity and distance in the z-space (Beta Variational Autoencoders didn’t have the same correlation in their latent space but showed comparatively good BLEU score, reconstruction error and perplexity). VAEs can also be used for generation of functional variants of peptides: Hawkins-Hooker et al., 2021, showed evidence  of conditional VAEs being able to increase solubility of luxA sequence from P. luminescens without losing core properties of the wild-type. Given the complexity of de novo design, diversification of already functional peptides can allow for a more reliable design of therapeutics. 

Most drugs target receptors  and channels (both are classes of proteins) on the surface of the cell or soluble (not part of a membrane) proteins. Given this, knowing the exact conformation of the target can be of much help when designing ligands. Moreover, proteins can serve as a therapeutics themselves. A huge milestone in deriving 3D structure from an amino acid sequence was achieved by DeepMind in protein structure annual competition, Critical Assessment of Protein Structure Prediction (CASP) \cite{noauthor_alphafold_nodate}. While their earlier architecture in CASP13, AlphaFold-1, relied on ResNet-like architecture to predict amino acid residue distances \cite{senior_improved_2020-1}, a major breakthrough was achieved with AlphaFold-2 in CASP14, when the team transitioned to transformer-like architecture \cite{noauthor_alphafold_nodate}. Previously, to predict the structure of a protein, one would have to undergo expensive and laborious X-ray crystallography or cryo-electron microscopy. Combination of generative methods with the knowledge of precise target shape can profoundly cut time for drug screening (Callaway, Ewen, 2020).  

\section{Drug response}
Apart from designing novel pharmaceutically active compounds, another important task is to be able to predict their efficiency. Ultimately, incorporating drug response directly as one of the optimization goals for generation of new molecules can lead to a more powerful drug design pipeline.
 
Recent research efforts have produced large-scale publicly available drug screening profiles (Table \ref{tab:cell_lines}). Most of them contain high-throughput screenings for anti-cancer drugs against cancer cell lines. These cell lines are most commonly immortalized human or animal cells which are grown outside an organism, on a Petri dish. Despite the tremendous value that cell lines provide to our understanding of disease progression and treatment \cite{mirabelli_cancer_2019} there are important drug response differences compared to full organisms. Therefore, there has been an ongoing effort to transfer findings from cell lines to actual application in humans where the amount of available data on drug responses is very limited.

The input features that are used for drug response prediction on cell lines consist of one or multiple of the following: expression levels or different genes, gene methylation status, and protein abundance. Less common are DNA mutation profiles (single nucleotide variations) and copy number variations. The nature of the problem -- expression profiles of hundreds or thousands of genes -- creates a high feature to sample ratio that complicates the prediction task.
 Initial approaches for drug response prediction included linear regression and simple clustering algorithms, SVM, and Random Forests. Traditional feature selection and dimensionality reduction approaches achieved some limited success \cite{adam_machine_2020, baptista_deep_2021-1}. While the task itself is not generative in nature, DGM models such as VAEs, conditional VAEs, and AAEs, have been used to perform dimensionality reduction, deconfounding or domain adaptation for better response prediction.

The first study to apply generative models to drug response predictions was described by Dincer et al. \cite{dincer_deepprofile_2018}. In this work, a VAE model has been used to learn relevant features from gene expression data of cancer cells treated with a drug. Compared to drug response data, there is a vast amount of unlabeled gene expression results which their model leveraged. First, the VAE was trained on gene expression data of AML patients and then, a much smaller sample size for which responses to various drugs were available were passed through the encoder. On the compressed data, a linear classifier was fit to predict a binary target if patients responded to therapy or not. This method improved performance over no feature reduction.

In another approach, a semi-supervised VAE called Dr.VAE divided the task of predicting drug response into two parts – first, it learns how a drug perturb gene expression of cells, and second, it relates gene expression changed to drug sensitivity \cite{rampasek_drvae_2019}. This means that instead of using only post-treatment gene expression data (as done above), the data is considered in pairs of pre- and post-treatment and there is a learnable function that describes this change. Dr.VAE aims to learn a joint distribution of gene expression pairs, latent representations, and a categorical variable that corresponds to the drug response. The marked improvement of this approach over baselines could be potentially even greater if combined with supplying the model with chemical properties of tested drugs. Combining gene expression data with embedded structure of chemical compounds was done by Kuenzi et al. \cite{kuenzi_predicting_2020} which used a simple feed-forward network.
 
Generative models have also been used for deconfounding in drug response prediction. As mentioned above, there is often a mismatch between drug efficiency tested in cell lines and in humans. When using predictive models, they can detect cell-line specific patterns. If an autoencoder is applied directly, it will mainly learn to encode these non-relevant signals as they are usually the main source of variation in the dataset. Adversarial Deconfounding Autoencoder (AD-AE) was used to extract only the relevant signal from this data (i.e., expression patterns of cancerous cells independent of their source) \cite{dincer_adversarial_2020}. This model is most similar to AAE, but the adversary network tries to predict a confounder, such as sex, biological age or origin of cell, from embeddings. This forces the latent space to contain only confounder-invariant representations. AD-AE was also better at domain adaptation to different datasets which indicates that it learned more meaningful cancer responses to drugs. Given that there is much more gene expression data on cancer cell lines than on tumor cells, AD-AE has also been applied for domain adaptation from cancer cells to tumors \cite{he_code-ae_2021-1}.
 
Finally, drug response can serve as a direct optimization target for generating new molecules. While conditional VAE (CVAE) has been used in the past for drug design with specific physicochemical properties (see section Drug Design; \cite{lim_molecular_2018}), Joo et al., used a conditional VAE used to generate new drug candidates with the desired property of inhibiting cancer cell growth \cite{joo_generative_2020}. The model takes in molecular fingerprints and a one-hot vector which indicates the binarized concentration of a drug needed to inhibit 50\% of cells in a population (GI50). During inference, a label with desired GI50 value is passed to the decoder along with a random sample from a latent space. Generated molecular fingerprints did indeed resemble similarity to those of FDA-approved drugs for breast cancer indicating that the model learned the relationship between the structural properties encoded in the fingerprints and efficiency of drugs. However, generating a molecular fingerprint is not the same as generating a full molecular structure. In the future, more direct approaches that generate full molecular graphs directly can be used with similar optimization targets.

\section{Challenges and Future Directions}
With exceptionally high failure rates and a large amount of unlabeled data, drug design and drug response prediction represent attractive areas for applying generative models. However, as emphasized by some authors in the field, reduction in cost and time for drug screening using AI often does not lead to improved decisions \cite{bender_dangers_2021-1}.
 
Generation of new pharmaceutically active compounds is challenging for several reasons. A lot of information is lost by using molecular representations. Many of the available data formats (such as SMILES or amino acid strings), when used alone, ignore the complexity of biological systems when trying to predict drug performance. This complexity includes but is not limited to secondary targets, drug metabolites and concentration-dependent effects.
 
Generating a molecular fingerprint of a drug or a SMILES string is far from generating a full molecular structure. Moreover, \textit{chiral} molecules (molecules that are not identical to their mirror images) often have very different properties. Drug receptors can recognize only one \textit{enantiomer} (one mirror image) but not the other. Sometimes, they can even have opposing properties with one being an inhibitor while the other being an activator of the same receptor \cite{weiskopf_drug_2002}. This information is completely lost in SMILES strings. Better molecular representations or modelling of full molecular graphs could therefore substantially advance the field in the upcoming years.
 
AlphaFold represents an excellent example of how emphasizing relevance of data (protein structure) can help solve the most complex biological problems \cite{senior_improved_2020-1}. In fact, Alphafold's transformer achieved state-of-art results in protein folding with just training 170,000 samples.
 
Another problem of some studies is a lack of proper benchmarking and evaluation compared to previously published work. Several papers provided only proof-of-the-concept examples of validity of their approaches. Having rigorous, standardized, and well-known comparison standards for drug design, similar to benchmarks for image generation on CIFAR-10 or ImageNet, would enable more transparent evaluation of newly proposed methods. Examples of proposed benchmarks include GuacaMol \cite{brown_guacamol_2019} or MOSES \cite{polykovskiy_molecular_2018}.
 
\begin{figure}[h!]
    \centering
    \includegraphics[width=\columnwidth]{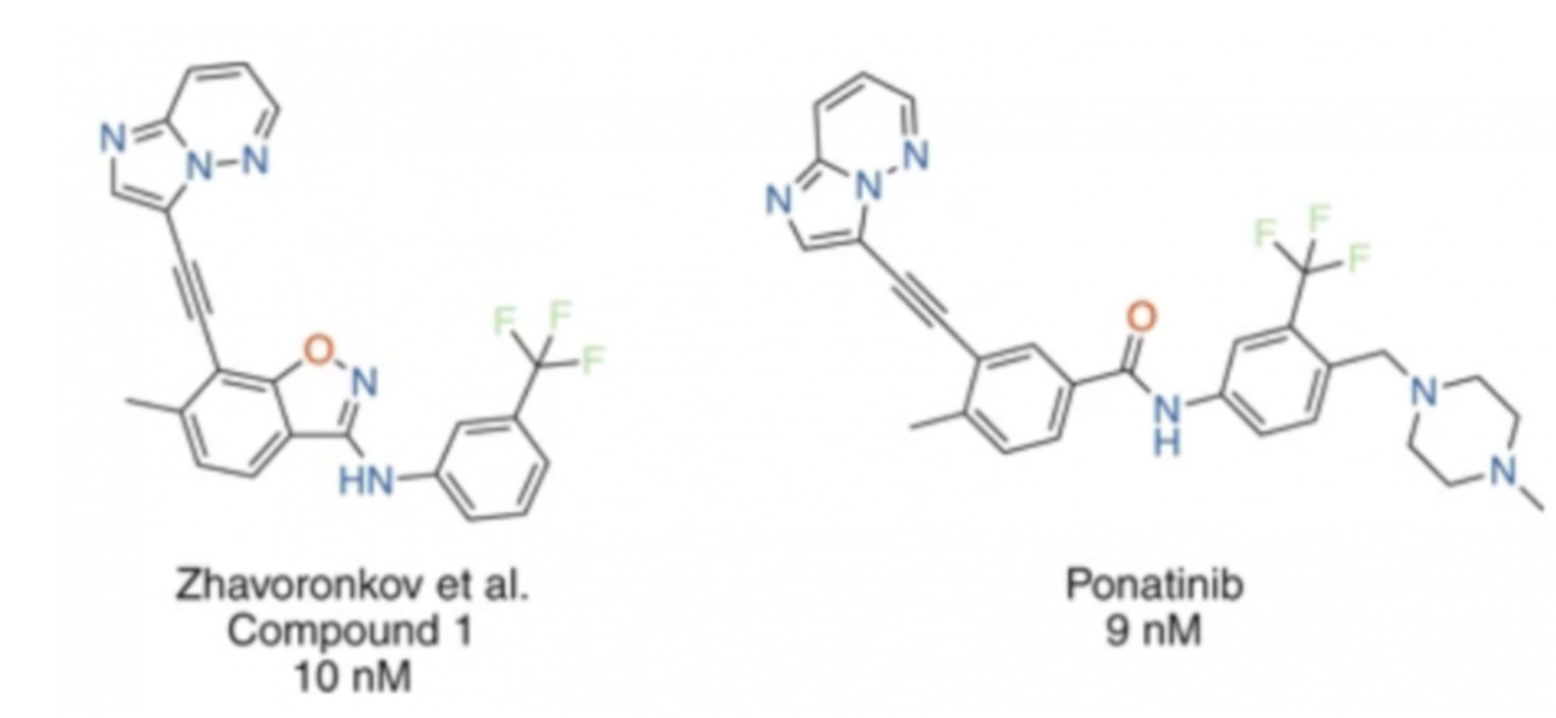}
    \caption{Compound 1 is DDR1 kinase inhibitor produced by the model and Ponatinib is a compound from dataset most similar to it. Adapted from }
    \label{fig:zhavor}
\end{figure}

Furthermore, the field of generation is particularly vulnerable to IP issues. For example, DDR1 kinase inhibitors recently produced by the generative pipeline of Insilico \cite{zhavoronkov_deep_2019} were claimed to be “the AlphaGO moment of drug design”. However, later a few researchers pointed out that the drug is almost identical to the one that appeared in the dataset of patented ligands they used, drug Iclusig (compound ponatinib) marketable by ARIAD Pharmaceuticals (Figure \ref{fig:zhavor}) \cite{walters_assessing_2020}. To prevent this from happening, the future research in de novo candidate generation should publish molecules from the training set that are most similar to the generated candidates.
This heavy reliance on data and previous research, however, points to one other problem. In recent years, there has been growing evidence on lack of reproducibility in biological research. Recent investigation into oncology drug design showed that two-thirds of published data on drug targets is not reproducible \cite{prinz_believe_2011}. Given wrong data, even state-of-the-art models won’t be able to generate meaningful candidates.

\section*{Conclusions}
In this paper, we present the current advances in the fields of drug design and drug response prediction using deep generative models. While there is a clear intuition for why these methods can be applied to drug discovery, most of the progress is yet to be made. Only a small fraction of research discussed in this review went beyond proof-of-concept and was used in industry pipelines. Generative models face limitations similar to previous approaches used in drug discovery and drug response prediction, (like QSAR), namely, little understanding of actual mechanism of action and risk of inaccurate predictions. Even though large databases of unlabeled molecules are attractive for unsupervised generations, to really increase success rate of drug discovery, future research should also make an emphasis on relevancy and quality of data used in the training.

\bibliographystyle{unsrt}
\bibliography{references}

\end{document}